\documentclass[conference]{IEEEtran}
\IEEEoverridecommandlockouts
\usepackage{cite}
\usepackage{amsmath,amssymb,amsfonts}
\usepackage{algorithmic}
\usepackage{graphicx}
\usepackage{textcomp}
\usepackage[table]{xcolor}
\usepackage{booktabs}
\usepackage{array}
\usepackage{url}

\definecolor{tablegray}{HTML}{EFEFEF}
\definecolor{tableblue}{HTML}{DCEBFF}

\def\BibTeX{{\rm B\kern-.05em{\sc i\kern-.025em b}\kern-.08em
    T\kern-.1667em\lower.7ex\hbox{E}\kern-.125emX}}

\begin{document}

\title{Collaborative Few-Step Distillation and Low-Bit Quantization for Wan2.2 Dual-Expert Video Diffusion Models%
\thanks{This technical report describes our submission to Sub-Challenge 1: W4A4 Quantization for Inference (HiF4 / MXFP4) of the GCC Low-Bit-width Large Model Quantization Challenge, an official challenge track of IEEE ICME 2026.}
}
\author{
\IEEEauthorblockN{Jinyang Du, Shenghao Jin, Ziqian Xu, Ruihao Gong, Shiqiao Gu, Yang Yong, Jinyang Guo, and Xianglong Liu}
\IEEEauthorblockA{
\{jinyangdu, shenghaojin, gongruihao, jinyangguo, xlliu\}@buaa.edu.cn, ziqianxu063@gmail.com\\
\{gushiqiao, yongyang\}@sensetime.com
}
}

\maketitle

\begin{abstract}
Large video diffusion models achieve strong visual quality but remain expensive to deploy because each sample requires many denoising steps and a large resident parameter footprint. This paper studies a deployment-oriented compression pipeline for Wan2.2-T2V-A14B by combining few-step distribution-matching distillation with low-bit quantization. The pipeline follows the model's dual-expert denoising route, calibrates the high-noise and low-noise branches separately, protects sensitive entrance layers, and uses HiF4-style low-bit representation to improve dynamic-range coverage. Quantization is calibrated on the distilled few-step student rather than on the original long-step trajectory, reducing activation-distribution mismatch during inference. The proposed co-design keeps the quantized model close to the same-step full-precision model and surpasses the original full-precision baseline at 8 and 20 steps on average. The 20-step setting gives the best quality-efficiency trade-off in the tested configurations.

\end{abstract}


\section{Overall Approach}
\subsection{Background and Objectives}
Diffusion models have become a dominant framework for image and video generation, but their iterative denoising process is expensive at deployment time \cite{ho2020ddpm, ho2022video}. Wan2.2 is an open video generation family whose A14B variant uses a dual-expert denoising architecture to improve capacity without activating all parameters at every step \cite{wan2025open}. This design improves generation quality but leaves two bottlenecks: long diffusion trajectories and large model memory. The goal of this work is to compress and accelerate an existing high-quality model through few-step distillation, low-bit quantization, and closed-loop video-quality evaluation.

The deployment target is a model that preserves subject identity, motion continuity, frame fidelity, and prompt alignment under limited GPU memory and latency budgets. Reducing only the number of sampling steps leaves the model expensive per step, whereas quantizing the original long-step model without matching the deployment trajectory may introduce activation-distribution mismatch and increase temporal error propagation. The core objective is therefore to first learn a shorter trajectory and then quantize under the activation distribution induced by that trajectory.

\subsection{End-to-End Technical Pipeline}
The pipeline is organized as
\textbf{model adaptation $\rightarrow$ few-step distillation $\rightarrow$ low-bit quantization $\rightarrow$ calibration construction $\rightarrow$ generation $\rightarrow$ evaluation}.
Implementation is built around Diffusers and LLMC \cite{diffusers, gong2024llmc}. First, the original model is adapted to a short denoising trajectory. Then, quantization calibration is performed under the same few-step deployment setting, so the quantizer observes the activation statistics it will meet at inference time. Finally, the full-precision and compressed models generate videos with matched prompts and sampling settings. The automatic evaluation stream computes VBench scores under a unified protocol, while the OpenS2V-based generation stream is used to produce submission videos for the human-scored competition metric.

Two principles guide the implementation. First, all calibration and replacement operations must follow the true high-noise and low-noise expert routes of Wan2.2 instead of treating the model as a dense trunk. Second, evaluation must compare models under matched generation conditions so that score changes can be attributed to the compression strategy rather than to prompt selection, sampling settings, or output organization. The pipeline is therefore both a compression method and a verification loop.

\section{Model, Data, and Evaluation Framework}
\subsection{Overview of Wan2.2}
Wan2.2-T2V-A14B decomposes denoising into a high-noise expert and a low-noise expert. The former handles global layout and motion planning, while the latter refines texture, lighting, and temporal detail. Although the full model contains about 27B parameters, only one roughly 14B expert is active at a given denoising step \cite{wan2025open}.

\begin{figure*}[t!]
\centering
\includegraphics[width=0.78\linewidth,height=1.20in,keepaspectratio]{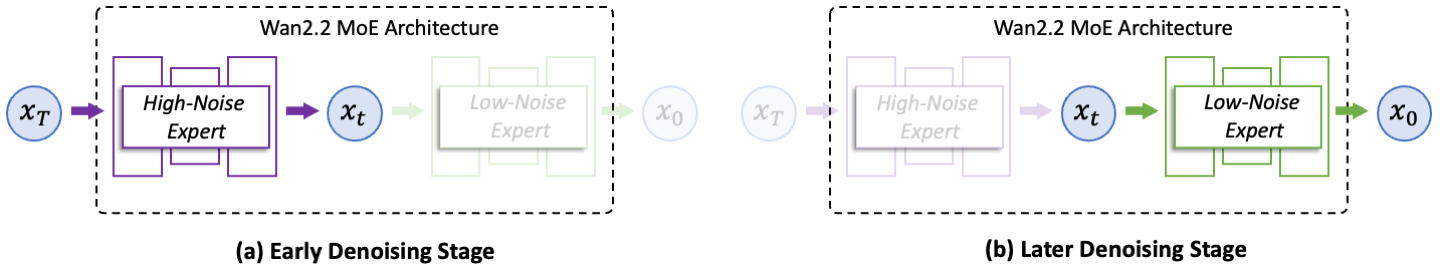}
\caption{Illustration of the Wan2.2 MoE architecture.}
\label{fig:wan_moe}
\end{figure*}

The branch switch makes compression different from single-trunk diffusion models. Calibration data, sensitive-layer protection, and layer replacement must follow the actual high-noise and low-noise execution paths; otherwise, the quantizer mixes incompatible activation distributions and amplifies low-bit error.

\subsection{Dataset Preparation and Preprocessing}
OpenS2V-5M is used as the prompt source for the OpenS2V-based submission stream and as a source for constructing calibration prompts, because it provides large-scale text-video data \cite{yuan2025opens2v}. The raw corpus is training-oriented, so it is converted into a compact inference stream. For each sample, the pipeline extracts the most subject-centered caption, preferring \texttt{face\_cap\_qwen} and falling back to \texttt{cap[0]} when necessary. Samples with low aesthetic quality or unstable motion are filtered out, and each retained item is mapped to \{\texttt{prompt}, \texttt{negative\_prompt}\}, with the negative prompt left empty when no explicit annotation is available. This preprocessing keeps calibration and evaluation prompts aligned with the subject-consistency demands of video generation.

The filtering stage is intentionally conservative. The calibration set should expose the model to ordinary subject and motion variation, but it should not be dominated by corrupted captions, low-quality videos, or extreme motion that makes evaluator scores noisy. Metadata fields that are irrelevant to text-to-video inference are removed, while the prompt text is kept close to the subject description rather than rewritten into a long synthetic instruction. This reduces the gap between calibration inputs and the prompts used by the VBench evaluation loop.

\subsection{VBench Evaluation Metrics}
Evaluation uses VBench, which decomposes video quality into interpretable dimensions \cite{huang2023vbench}. We focus on five metrics: Subject Consistency, Aesthetic Quality, Imaging Quality, Overall Consistency, and Motion Smoothness. Subject Consistency uses DINO features to compare adjacent frames and first-frame anchors \cite{caron2021emerging}. Aesthetic Quality uses CLIP features and a LAION aesthetic predictor \cite{radford2021learning, schuhmann2022laion}; Imaging Quality uses MUSIQ \cite{ke2021musiq}; Overall Consistency compares video and text features through ViCLIP/InternVideo representations \cite{wang2023internvid}; and Motion Smoothness uses AMT-based frame interpolation \cite{li2023amt}. Together, these metrics cover subject identity, perceptual quality, frame fidelity, prompt alignment, and temporal continuity.

\begin{table}[t!]
\centering
\caption{VBench dimensions used in the experiments.}
\label{tab:metrics}
\scriptsize
\setlength{\tabcolsep}{3pt}
\renewcommand{\arraystretch}{1.08}
\resizebox{\linewidth}{!}{%
\begin{tabular}{lll}
\toprule
\textbf{Metric} & \textbf{Evaluator} & \textbf{Main sensitivity} \\
\midrule
SC & DINO features & subject identity over time \\
AQ & CLIP + LAION predictor & composition and visual appeal \\
IQ & MUSIQ & blur, artifacts, and detail fidelity \\
OC & ViCLIP/InternVideo & text-video semantic alignment \\
MS & AMT interpolation & temporal continuity and motion jitter \\
\bottomrule
\end{tabular}%
}
\end{table}

The selected dimensions are chosen because they correspond to the most visible failure modes after compression. A quantized video model may still produce sharp individual frames while the subject slowly drifts, or it may preserve the main subject while reducing prompt faithfulness. Reporting these dimensions separately avoids hiding such trade-offs behind a single average. The average score is still useful as a compact summary, but per-metric values are needed to determine whether the compression pipeline is deployment-ready.

\subsection{HiFloat4 Number Format}
HiF4 is a 4-bit block floating-point format designed for low-bit inference \cite{luo2026hifloat4}. Each unit stores 64 signed 4-bit values plus 32 bits of shared scaling metadata, giving an average cost of 4.5 bits per value. The metadata contains a three-level scaling hierarchy:
\[
\underset{\text{Level-1:\ E6M2}}{\underbrace{8\text{bits}}}
+ \underset{\text{Level-2:\ E1\_8}}{\underbrace{8 \times 1\text{bits}}}
+ \underset{\text{Level-3:\ E1\_16}}{\underbrace{16 \times 1\text{bits}}}
= 32\text{bits}.
\]

\begin{figure}[t!]
\centering
\includegraphics[width=0.82\linewidth,height=1.10in,keepaspectratio]{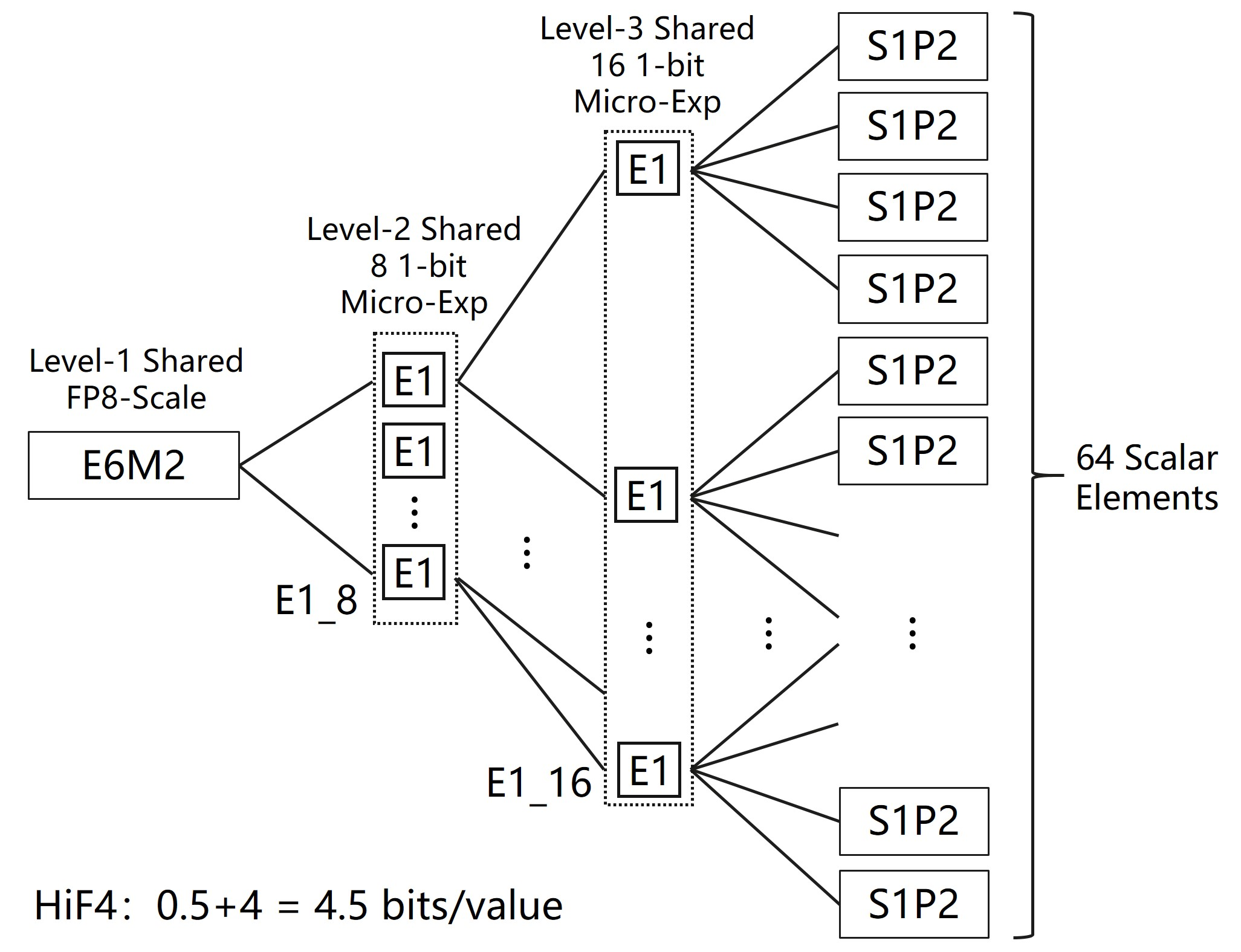}
\caption{Three-level scaling hierarchy in HiF4.}
\label{fig:hif4}
\end{figure}

\textbf{Global and local scaling.}
The first level stores an E6M2 base scale,
\[
X_{\text{E6M2}} = 2^{E-48} \times 1.M.
\]
With exponent bias 48, the E6M2 base scale provides the coarse exponent range. 
The second and third levels add binary micro-exponents for sub-groups, while the 
4-bit in-block value uses S1P2 signed amplitude. The reconstructed value is
\[
V_i
= \text{E6M2}
\times 2^{\{E1_{8}\}_{\lceil i/8\rceil} + \{E1_{16}\}_{\lceil i/4\rceil}}
\times \{S1P2\}_i.
\]
Following the HiF4 specification, the combined E6M2--E1--E1--S1P2 hierarchy 
provides a global representable range from approximately \(2^{-50}\) to \(2^{18}\). 
Let \(v_{\min}^{+}\) and \(v_{\max}\) denote the smallest positive and largest 
representable magnitudes. The number of covered binades can be written as
\[
R_{\mathrm{global}}
=
\left\lfloor \log_2 v_{\max} \right\rfloor
-
\left\lfloor \log_2 v_{\min}^{+} \right\rfloor
+ 1
=
18 - (-50) + 1
=
69,
\]
while the micro-exponents and S1P2 values provide a local range of
\[
\log_2\!\left(\frac{2^{1+1}\times1.75}{0.25}\right)
\approx 4.81.
\]
This hierarchy is useful for Wan2.2 because high-noise and low-noise experts produce different activation ranges, and temporal-spatial tokens are strongly heterogeneous. A wider global range helps avoid overflow, while local sub-group scaling preserves low-amplitude detail that affects texture, subject identity, and motion smoothness. Compared with a narrow 4-bit floating format that must rely more heavily on external scaling, HiF4 keeps more range control inside the block itself, which simplifies deployment when activation statistics change across denoising phases.

The choice of HiF4 is therefore tied to the structure of the target model rather than only to a nominal bitwidth. In early denoising, activations are influenced by high noise levels and global layout formation; in late denoising, the same network family must represent fine detail and small temporal corrections. A format with insufficient dynamic range tends to overprotect large values and underrepresent small but perceptually important changes. HiF4's hierarchical scale design is intended to reduce this conflict under a fixed low-bit budget.

\section{Quantization and Few-Step Distillation Design}
\subsection{Quantization Method Selection and Model Adaptation}
\subsubsection{Quantization Method Selection and Basic Formulation}

We use a lightweight rounding-based post-training quantization method and improve it through granularity design, branch-wise calibration, and sensitive-layer protection. Prior PTQ methods, such as GPTQ and SmoothQuant, show that calibration design is critical for large transformer-style models \cite{frantar2023gptq, xiao2023smoothquant}; here, the same principle is adapted to video diffusion.

In generic form, quantization can be written as
\[
\hat{x}
= s \cdot \operatorname{clip}\!\left(
\operatorname{round}\!\left(\frac{x}{s}\right),
q_{\min},
q_{\max}
\right),
\]
where $x$ denotes the tensor to be quantized. The scale can be selected by minimizing
\[
\min_s \mathbb{E}\!\left[(x-\hat{x})^2\right].
\]
In the implementation, weights are quantized per channel and activations per token. Per-channel scaling reduces weight-outlier effects, while per-token scaling better follows the prompt-, frame-, and stage-dependent activation variation of video diffusion features.

\begin{table}[t!]
\centering
\caption{Quantization configuration used in the final pipeline.}
\label{tab:quant_config}
\scriptsize
\setlength{\tabcolsep}{3pt}
\renewcommand{\arraystretch}{1.10}
\resizebox{\linewidth}{!}{%
\begin{tabular}{lll}
\toprule
\textbf{Part} & \textbf{Choice} & \textbf{Purpose} \\
\midrule
Weights & per-channel low-bit scaling & reduce channel outlier error \\
Activations & per-token calibration & track prompt- and frame-level variation \\
Branches & separate expert passes & preserve high-/low-noise statistics \\
Entrance layers & protected or enhanced calibration & suppress early error propagation \\
Format & HiF4-style BFP & improve dynamic-range coverage \\
\bottomrule
\end{tabular}%
}
\end{table}

\subsubsection{Quantization Adaptation for the Wan2.2 Dual-Expert Structure}

The dual-expert routing at time step $t$ can be written as
\[
f_t(z)=
\begin{cases}
E_{\text{high}}(z,t), & t \geq t_{\text{moe}},\\
E_{\text{low}}(z,t), & t < t_{\text{moe}}.
\end{cases}
\]
The two branches are calibrated and replaced separately. Inputs to the first block of each expert are captured along the true inference route, after which each branch is processed layer by layer. This converts quantization from one traversal over a presumed dense trunk into two coordinated passes over the real MoE paths.

Concretely, let $\mathcal{D}_{\text{cal}}$ be the calibration prompts and let $\mathcal{A}^{h}_{\ell}$ and $\mathcal{A}^{l}_{\ell}$ be the activation sets captured for layer $\ell$ in the high-noise and low-noise experts. The calibration objective is not computed on a pooled activation set, but on branch-specific statistics:
\[
\begin{aligned}
\min_{Q_h,Q_l}\sum_{\ell} \big(&
\mathbb{E}_{a \in \mathcal{A}^{h}_{\ell}}
\left[\|a-Q_h(a)\|_2^2\right] \\
&+
\mathbb{E}_{a \in \mathcal{A}^{l}_{\ell}}
\left[\|a-Q_l(a)\|_2^2\right]\big).
\end{aligned}
\]
This formulation keeps the quantizer aligned with the expert that will actually process a given stage: high-noise activations carry global structure and motion layout, while low-noise activations are more tied to texture, boundary detail, and local temporal repair. After each layer is quantized, its output is propagated before calibrating the next layer, so later layers see realistic upstream low-bit errors.

\subsubsection{Entrance-Layer Protection and Enhanced Calibration}

Very early errors are repeatedly propagated through denoising, so the first blocks and a small set of entrance layers are kept in higher precision or calibrated more aggressively. Let $\mathcal{K}$ denote the protected set. The effective quantized model is
\[
\tilde{f}_t =
\tilde{E}_{\text{high/low}}\!\left(\mathcal{K}_{\text{fp}}, \mathcal{K}_{\text{q}}\right),
\]
where $\mathcal{K}_{\text{fp}}$ contains protected layers and $\mathcal{K}_{\text{q}}$ contains low-bit layers. This design concentrates precision where error propagation is most harmful while still obtaining most of the memory benefit from low-bit weights and activations.

The protected set remains small: entrance and early routing-sensitive layers receive priority because they shape the latent state before stable subject layout is established, whereas later-layer errors are usually more local and partly absorbed by refinement.

\subsection{Integrating Few-Step Distillation}
\subsubsection{Goals and Advantages of Few-Step Distillation}

Few-step distillation reduces the number of denoising evaluations while relearning the short trajectory rather than merely truncating the sampler. This follows a line of fast diffusion sampling work, including progressive distillation and consistency models \cite{salimans2022progressive, song2023consistency}. Let $C_{\text{step}}$ be the cost of one denoising step and $p_S(x;\theta)$ the student generation distribution. We compare 4-, 8-, 20-, and 40-step settings, with the 20-step distilled model selected as the main trade-off point.

Unlike truncation, distillation optimizes intermediate states for the reduced number of evaluations. This makes the student a better quantization target because the calibration distribution is closer to deployment.

\subsubsection{Distillation Optimization via Distribution Matching and Stage-Aware Modeling}

DMD aligns the student with the teacher distribution rather than with a fixed one-to-one sampling path \cite{yin2024dmd}. For Wan2.2, the target should also respect the high-noise and low-noise phases. The early phase establishes layout, subject positions, and motion; the late phase refines texture, boundary quality, and local temporal coherence.

This motivates a stage-aware objective such as
\[
\mathcal{L}_{\text{distill}}
= \lambda_1 \mathcal{L}_{\text{high}}
+ \lambda_2 \mathcal{L}_{\text{low}}.
\]
Phased DMD extends this idea by matching score behavior within subintervals and is especially suitable for MoE-style video denoising \cite{fan2025phased}. It makes the student more stable than a naive short-step sampler and improves its suitability for later low-bit compression.

For deployment, the distilled student is treated as a new inference model with its own numerical behavior. A short trajectory changes which denoising states are visited, how long each expert remains active, and how activation magnitudes evolve. This is why the experiments compare several step counts instead of assuming that the fewest steps are automatically best.

\subsubsection{Co-Design of Few-Step Distillation and Quantization}

At a given time step $t$, the activation distribution of the student generally differs from that of the original high-step reference:
\[
p_t^{\text{student}}(a) \neq p_t^{\text{ref}}(a).
\]
This mismatch is structural. If calibration still uses the long-step reference statistics, the quantizer sees a distribution different from deployment, and low-bit error is amplified. Calibration is therefore performed on the distilled student under the true few-step sampling schedule, with branch-wise statistics preserved.

This co-design offers two direct benefits. First, few-step distillation already reduces the sampling length, so the total inference cost can be approximated as
\[
C_{\text{total}} \approx N \cdot C_{\text{step}} + C_{\text{overhead}},
\]
where $N$ is the number of sampling steps. Second, calibrating directly on student trajectories reduces distribution mismatch and makes the quantized model more likely to preserve both global structure and local detail under low-bit deployment.

Thus, the distillation constraint preserves the teacher distribution under fewer denoising states, while the quantization constraint preserves the student's branch-specific activations under those states. This order is more stable than quantizing the original long-step model first because it avoids calibrating a distribution that will not be used after acceleration.

\subsection{Engineering Implementation and Key Challenges}
\subsubsection{Memory Analysis and OOM Control}

Peak memory is approximated as
\[
M_{\text{peak}} \approx M_{\text{param}} + M_{\text{act}} + M_{\text{aux}},
\]
where $M_{\text{param}}$ is the memory occupied by resident parameters, $M_{\text{act}}$ is the activation footprint, and $M_{\text{aux}}$ is the overhead of auxiliary buffers.

To avoid out-of-memory failures, calibration and replacement are performed layer by layer. Only the current block and necessary buffers are resident during statistics collection and adjustment. Auxiliary distillation modules are also loaded on demand: they move to the GPU only for the stage that needs them and return to the CPU afterward. This trades limited CPU--GPU communication for a much lower resident memory baseline.

This memory policy also improves reproducibility. A full-network calibration run is sensitive to small changes in cache state and auxiliary tensors, whereas layer-wise processing fixes the active memory set for each block. For the dual-expert model, the same rule is applied independently to the high-noise and low-noise branches: capture inputs, quantize the current block, validate the replacement, release temporary buffers, and then continue to the next block. The process is slower than a fully resident calibration pass, but it is more reliable on a single 80GB GPU and avoids changing the algorithmic design to fit the hardware.

\subsubsection{Closed-Loop Evaluation and System Verification}

Verification uses batch generation rather than isolated qualitative examples. The full-precision few-step model and the quantized model share prompts, resolution, output format, and sampling schedule. Scores are then aggregated over the selected VBench dimensions, so quality differences can be traced to subject stability, visual quality, semantic alignment, or temporal motion.

A run is accepted only when generation and evaluation are both completed under the same prompt index. This prevents missing videos or mismatched filenames from silently changing the metric denominator. The resulting evaluation loop is therefore useful not only for reporting final numbers, but also for locating failure modes during development: a drop in SC or MS points to temporal propagation, a drop in IQ points to local frame fidelity, and a drop in OC points to semantic drift between prompts and generated videos.

\section{Experimental Results and Analysis}
\subsection{Experimental Setup}
\subsubsection{Hardware Environment and Resource Consumption}

Experiments were run on Ubuntu 22.04.3 with dual Intel Xeon Platinum 8462Y+ CPUs, 2.0 TiB memory, and one NVIDIA H100 80GB GPU using CUDA 12.4. Inference, quantization, and evaluation were separated into dedicated Conda environments to avoid dependency conflicts: Wan2.2 inference used PyTorch 2.5.1 and Diffusers 0.37.0; LLMC quantization used PyTorch 2.8.0; and VBench evaluation used PyTorch 2.4.1 with VBench 0.1.5.

The separation of environments is not merely organizational. The inference stack must match the model loader and generation scheduler, the quantization stack must support low-bit replacement and layer-wise calibration, and the evaluation stack must load the pretrained metric models used by VBench. Keeping these stages isolated reduces version conflicts and makes failed runs easier to trace.

\subsubsection{Baselines and Controlled Variables}

The experiments compare three types of systems:

\begin{itemize}
\item the original full-precision baseline model,
\item the few-step distilled full-precision model, and
\item the corresponding few-step quantized model.
\end{itemize}

To ensure fair comparisons, the following variables are controlled throughout the experiments:

\begin{itemize}
\item sampling steps: 4, 8, 20, and 40;
\item evaluation dataset: VBench;
\item evaluation metrics: SC, AQ, IQ, OC, and MS.
\end{itemize}

The number of sampling steps measures the efficiency gain brought by distillation, while the selected metrics correspond to the core visual, semantic, and temporal dimensions emphasized by VBench.

\subsubsection{Data Construction and Evaluation Workflow}

The quantitative workflow starts from the standard VBench prompt metadata. Prompts belonging to the five selected dimensions are filtered, generated videos are stored by dimension, and the VBench evaluator builds a run-time JSON index that maps prompts to concrete video files. Each metric module then reads the same index and computes its score with the corresponding pretrained evaluator. The generated outputs are not manually selected; all located samples for a dimension enter the corresponding score. This design keeps the generation protocol identical across baselines and makes the final average score a controlled comparison rather than a collection of unrelated examples.

The five dimensions are complementary. SC and MS are most sensitive to temporal error accumulation, AQ and IQ focus on visual appearance and frame fidelity, and OC measures text-video semantic alignment. This separation is important when interpreting quantization results: a model can improve frame sharpness while still damaging motion, or preserve prompt alignment while weakening subject identity. The final comparison therefore reports both average scores and per-metric scores for the representative setting.

All baselines are evaluated with the same prompt index and output naming convention. The original full-precision model serves as the deployment-quality reference, the same-step full-precision model isolates the effect of few-step distillation, and the quantized model measures the additional effect of low-bit conversion. This three-way comparison is necessary because a direct comparison between the original model and the final quantized model would mix step reduction and quantization into a single number.

\subsection{Core Experimental Results}
After few-step distillation is introduced, the number of denoising steps becomes the main efficiency variable. Table~\ref{tab:steps_compare} compares the original full-precision model, the same-step full-precision model, and the quantized model. Repeated runs are averaged where available, and the 40-step row is kept as a long-trajectory reference.

\begin{table}[t!]
\centering
\caption{Average-score comparison across sampling-step settings.}
\label{tab:steps_compare}
\scriptsize
\setlength{\tabcolsep}{4pt}
\renewcommand{\arraystretch}{1.08}
\resizebox{\linewidth}{!}{%
\begin{tabular}{cccc}
\toprule
\textbf{Steps} & \textbf{Original FP} & \textbf{Same-Step FP} & \textbf{Quantized} \\
\midrule
4  & 0.6976 & 0.6795 & 0.6802 \\
8  & 0.6976 & 0.6991 & 0.7002 \\
20 & 0.6976 & 0.7051 & 0.7074 \\
40 & 0.6993 & 0.7084 & 0.7077 \\
\bottomrule
\end{tabular}%
}
\end{table}

The 4-step model already keeps quantization error small relative to the same-step full-precision model, but the short trajectory still trails the original baseline. At 8 steps, the quantized model reaches 0.7002 and begins to exceed the original full-precision score. At 20 steps, it reaches 0.7074, outperforming both the original model and the same-step full-precision model. The 40-step result is similar to the 20-step result, indicating diminishing returns from further increasing the sampling length. Thus, the 20-step configuration gives the strongest quality--efficiency balance in this study.

The same-step comparison is the key control: the quantized model remains close to or slightly above the same-step full-precision model, meaning that low-bit conversion does not introduce an additional quality penalty after the student trajectory is fixed.

\begin{table}[t!]
\centering
\caption{Per-metric comparison for the representative 20-step setting.}
\label{tab:best_case}
\scriptsize
\setlength{\tabcolsep}{3pt}
\renewcommand{\arraystretch}{1.08}
\resizebox{\linewidth}{!}{%
\begin{tabular}{lcccccc}
\toprule
\textbf{Method} & \textbf{SC} & \textbf{AQ} & \textbf{IQ} & \textbf{OC} & \textbf{MS} & \textbf{Avg} \\
\midrule
\textbf{Original FP} & 0.9495 & 0.6474 & 0.6479 & 0.2524 & 0.9819 & 0.6958 \\
\rowcolor{tablegray} \textbf{Same-Step FP} & 0.9597 & 0.6553 & 0.6828 & 0.2422 & 0.9821 & 0.7044 \\
\rowcolor{tableblue} \textbf{Quantized} & 0.9629 & 0.6571 & 0.6872 & 0.2434 & 0.9825 & 0.7066 \\
\bottomrule
\end{tabular}%
}
\end{table}

Table~\ref{tab:best_case} confirms that the 20-step gain is not caused by a single metric. The quantized model improves Subject Consistency, Aesthetic Quality, Imaging Quality, and Motion Smoothness over the original full-precision baseline, while Overall Consistency remains close to the same-step model. This suggests that distribution-aligned calibration recovers most low-bit loss without sacrificing the temporal metrics that are usually most fragile in video generation.

\subsection{Error Analysis}
\subsubsection{Metric-Specific Error Sensitivity}

Quantization does not affect all VBench dimensions equally. Single-frame perturbations mainly influence Imaging Quality and Aesthetic Quality, while errors that accumulate across denoising steps and video frames first appear in Subject Consistency and Motion Smoothness. This explains why frame-level and temporal metrics should be interpreted separately when evaluating low-bit video diffusion models.

This asymmetry is characteristic of video generation. A small numerical perturbation may be visually tolerable in one isolated frame, but the same perturbation can alter the latent trajectory and become visible as subject drift, flicker, or unstable motion several frames later. Therefore, a quantization method that is acceptable for image generation is not automatically sufficient for video generation. The method must explicitly control where errors enter the trajectory and how they propagate across time.

\subsubsection{First-Frame and Entrance-Layer Effects}

Subject Consistency compares frames not only with their neighbors but also with the first frame, so an unstable first frame becomes a repeated reference error. Entrance layers and early high-noise blocks are therefore disproportionately important: if they distort subject layout or motion initialization, later refinement cannot fully repair the sequence. Protecting these layers and calibrating them more carefully controls the main error-propagation path, which is consistent with the improved temporal metrics in the final quantized model.

The high-noise expert is especially sensitive because it determines coarse composition and motion direction before detailed refinement begins. Once this stage places the subject incorrectly or creates an unstable motion plan, the low-noise expert tends to refine the wrong structure rather than replace it. Conversely, errors in late low-noise layers are more likely to appear as texture loss, blurred edges, or local jitter. This division of failure modes supports the branch-wise calibration strategy used in the proposed pipeline.

\subsection{Summary of Advantages}
\subsubsection{Quality and Efficiency}

The main advantage of the proposed pipeline is that it compresses both axes of inference cost. Few-step distillation reduces the number of denoising evaluations, and low-bit quantization reduces the cost and memory footprint of each evaluation. The distribution-aligned few-step quantized model approaches the same-step full-precision model at 4 steps and surpasses the original full-precision baseline at 8 and 20 steps.

The results also show that the best operating point is not the shortest trajectory. The 4-step setting gives strong acceleration but still leaves a visible gap to the original full-precision baseline. The 20-step setting provides enough denoising capacity for the distilled model to recover temporal and visual quality, while still requiring far fewer steps than a conventional long sampler. This makes it the most balanced configuration among the tested options.

\subsubsection{Deployment Value}

The 20-step result shows that careful co-design can turn quantization from a lossy afterthought into part of the acceleration strategy. Branch-wise calibration respects Wan2.2's dual-expert structure, protected entrance layers reduce temporal error propagation, and HiF4-style representation improves numerical coverage under a fixed bit budget. Together, these choices provide a practical path for deploying high-quality video diffusion models under limited GPU memory and latency budgets.

There are still limitations. The current pipeline focuses on post-training quantization and deployment-time evaluation rather than retraining the full generator. It therefore depends on the quality of the distilled student and on how well the calibration prompts cover deployment prompts. Future work could add prompt-aware calibration, mixed precision guided by layer sensitivity, or distillation objectives that explicitly weight semantic alignment and long-horizon temporal consistency.

\subsubsection{Practical Takeaways}

For practical deployment, the compression target should be selected from the joint quality--efficiency curve rather than from bitwidth or step count alone. The 4-step setting is attractive when latency is the only priority, but it leaves less room for correcting layout and temporal artifacts. The 20-step setting is more suitable when stable subject identity and reliable motion are required, making it a reasonable default for interactive generation, preview rendering, and resource-constrained batch generation.

The calibration procedure should also be treated as part of the deployment specification. If calibration prompts are too narrow, the quantizer may fail on prompts with unusual subject scale, fast motion, or complex backgrounds; if they are too noisy, the collected activation ranges become less representative of high-quality generation. Filtering OpenS2V prompts by subject clarity and quality indicators helps keep the calibration set close to the intended use case, while larger deployments should refresh calibration data when the prompt distribution changes.

Full uniform quantization is not always the best engineering choice. Keeping a few entrance and routing-sensitive layers in higher precision costs little compared with the total model size, but protects stages where errors have the largest downstream effect. This is especially important for dual-expert video diffusion models, where early high-noise errors influence the entire sequence and late low-noise errors mainly appear as local texture or boundary artifacts.

Finally, the evaluation loop is useful beyond final reporting. During development, metric-specific failures give direct debugging signals: Subject Consistency points to identity drift or first-frame instability, Motion Smoothness reflects temporal discontinuity, Imaging Quality exposes frame-level artifacts, and Overall Consistency indicates prompt-video semantic mismatch. This makes the pipeline iterative: adjust calibration, protected layers, or step count, then rerun the same VBench protocol to verify targeted improvements without damaging other dimensions.

\bibliographystyle{IEEEtran}
\bibliography{references}

@article{wan2025open,
  title = {Wan: Open and Advanced Large-Scale Video Generative Models},
  author = {{Wan Team} and Wang, Ang and Ai, Baole and Wen, Bin and Mao, Chaojie and Xie, Chen-Wei and Chen, Di and Yu, Feiwu and Zhao, Haiming and Yang, Jianxiao and others},
  journal = {arXiv preprint arXiv:2503.20314},
  year = {2025},
  url = {https://arxiv.org/abs/2503.20314}
}

@inproceedings{ho2020ddpm,
  title = {Denoising Diffusion Probabilistic Models},
  author = {Ho, Jonathan and Jain, Ajay and Abbeel, Pieter},
  booktitle = {Advances in Neural Information Processing Systems},
  volume = {33},
  pages = {6840--6851},
  year = {2020}
}

@article{ho2022video,
  title = {Video Diffusion Models},
  author = {Ho, Jonathan and Salimans, Tim and Gritsenko, Alexey and Chan, William and Norouzi, Mohammad and Fleet, David J.},
  journal = {arXiv preprint arXiv:2204.03458},
  year = {2022},
  url = {https://arxiv.org/abs/2204.03458}
}

@misc{diffusers,
  title = {Diffusers: State-of-the-art Diffusion Models},
  author = {von Platen, Patrick and Patil, Suraj and Cuenca, Pedro and Lambert, Nathan and Rasul, Kashif and Davaadorj, Mishig and Nair, Deepak and Paul, Sayak and Berman, William and Xu, Yiyi and Liu, Steven and Wolf, Thomas},
  year = {2022},
  howpublished = {\url{https://github.com/huggingface/diffusers}}
}

@article{gong2024llmc,
  title = {{LLMC}: Benchmarking Large Language Model Quantization with a Versatile Compression Toolkit},
  author = {Gong, Ruihao and Yong, Yang and Gu, Shiqiao and Huang, Yushi and Lv, Chengtao and Zhang, Yunchen and Liu, Xianglong and Tao, Dacheng},
  journal = {arXiv preprint arXiv:2405.06001},
  year = {2024},
  url = {https://arxiv.org/abs/2405.06001}
}

@inproceedings{yin2024dmd,
  title = {One-step Diffusion with Distribution Matching Distillation},
  author = {Yin, Tianwei and Gharbi, Michael and Zhang, Richard and Shechtman, Eli and Durand, Fredo and Freeman, William T. and Park, Taesung},
  booktitle = {Proceedings of the IEEE/CVF Conference on Computer Vision and Pattern Recognition},
  year = {2024},
  url = {https://arxiv.org/abs/2311.18828}
}

@inproceedings{salimans2022progressive,
  title = {Progressive Distillation for Fast Sampling of Diffusion Models},
  author = {Salimans, Tim and Ho, Jonathan},
  booktitle = {International Conference on Learning Representations},
  year = {2022},
  url = {https://arxiv.org/abs/2202.00512}
}

@inproceedings{song2023consistency,
  title = {Consistency Models},
  author = {Song, Yang and Dhariwal, Prafulla and Chen, Mark and Sutskever, Ilya},
  booktitle = {Proceedings of the 40th International Conference on Machine Learning},
  volume = {202},
  pages = {32211--32252},
  year = {2023},
  url = {https://proceedings.mlr.press/v202/song23a.html}
}

@article{fan2025phased,
  title = {Phased {DMD}: Few-step Distribution Matching Distillation via Score Matching within Subintervals},
  author = {Fan, Xiangyu and Qiu, Zesong and Wu, Zhuguanyu and Wang, Fanzhou and Lin, Zhiqian and Ren, Tianxiang and Lin, Dahua and Gong, Ruihao and Yang, Lei},
  journal = {arXiv preprint arXiv:2510.27684},
  year = {2025},
  url = {https://arxiv.org/abs/2510.27684}
}

@article{huang2023vbench,
  title = {{VBench}: Comprehensive Benchmark Suite for Video Generative Models},
  author = {Huang, Ziqi and He, Yinan and Yu, Jiashuo and Zhang, Fan and Si, Chenyang and Jiang, Yuming and Zhang, Yuanhan and Wu, Tianxing and Jin, Qingyang and Chanpaisit, Nattapol and Wang, Yaohui and Chen, Xinyuan and Wang, Limin and Lin, Dahua and Qiao, Yu and Liu, Ziwei},
  journal = {arXiv preprint arXiv:2311.17982},
  year = {2023},
  url = {https://arxiv.org/abs/2311.17982}
}

@inproceedings{frantar2023gptq,
  title = {{GPTQ}: Accurate Post-Training Quantization for Generative Pre-trained Transformers},
  author = {Frantar, Elias and Ashkboos, Saleh and Hoefler, Torsten and Alistarh, Dan},
  booktitle = {International Conference on Learning Representations},
  year = {2023},
  url = {https://arxiv.org/abs/2210.17323}
}

@inproceedings{xiao2023smoothquant,
  title = {{SmoothQuant}: Accurate and Efficient Post-Training Quantization for Large Language Models},
  author = {Xiao, Guangxuan and Lin, Ji and Seznec, Mickael and Wu, Hao and Demouth, Julien and Han, Song},
  booktitle = {Proceedings of the 40th International Conference on Machine Learning},
  volume = {202},
  pages = {38087--38099},
  year = {2023},
  url = {https://arxiv.org/abs/2211.10438}
}

@inproceedings{caron2021emerging,
  title = {Emerging Properties in Self-Supervised Vision Transformers},
  author = {Caron, Mathilde and Touvron, Hugo and Misra, Ishan and Jegou, Herve and Mairal, Julien and Bojanowski, Piotr and Joulin, Armand},
  booktitle = {Proceedings of the IEEE/CVF International Conference on Computer Vision},
  pages = {9650--9660},
  year = {2021}
}

@inproceedings{radford2021learning,
  title = {Learning Transferable Visual Models From Natural Language Supervision},
  author = {Radford, Alec and Kim, Jong Wook and Hallacy, Chris and Ramesh, Aditya and Goh, Gabriel and Agarwal, Sandhini and Sastry, Girish and Askell, Amanda and Mishkin, Pamela and Clark, Jack and Krueger, Gretchen and Sutskever, Ilya},
  booktitle = {Proceedings of the International Conference on Machine Learning},
  pages = {8748--8763},
  year = {2021}
}

@inproceedings{ke2021musiq,
  title = {{MUSIQ}: Multi-Scale Image Quality Transformer},
  author = {Ke, Junjie and Wang, Qifei and Wang, Yilin and Milanfar, Peyman and Yang, Feng},
  booktitle = {Proceedings of the IEEE/CVF International Conference on Computer Vision},
  pages = {5148--5157},
  year = {2021},
  doi = {10.1109/ICCV48922.2021.00510}
}

@article{wang2023internvid,
  title = {{InternVid}: A Large-scale Video-Text Dataset for Multimodal Understanding and Generation},
  author = {Wang, Yi and He, Yinan and Li, Yizhuo and Li, Kunchang and Yu, Jiashuo and Ma, Xin and Chen, Xinyuan and Wang, Yaohui and Luo, Ping and Liu, Ziwei and Wang, Yali and Wang, Limin and Qiao, Yu},
  journal = {arXiv preprint arXiv:2307.06942},
  year = {2023},
  url = {https://arxiv.org/abs/2307.06942}
}

@article{schuhmann2022laion,
  title = {{LAION-5B}: An Open Large-Scale Dataset for Training Next Generation Image-Text Models},
  author = {Schuhmann, Christoph and Beaumont, Romain and Vencu, Richard and Gordon, Cade and Wightman, Ross and Cherti, Mehdi and Coombes, Theo and Katta, Aarush and Mullis, Clayton and Wortsman, Mitchell and Schramowski, Patrick and Kundurthy, Srivatsa and Crowson, Katherine and Schmidt, Ludwig and Kaczmarczyk, Robert and Jitsev, Jenia},
  journal = {Advances in Neural Information Processing Systems},
  volume = {35},
  pages = {25278--25294},
  year = {2022}
}

@inproceedings{li2023amt,
  title = {{AMT}: All-Pairs Multi-Field Transforms for Efficient Frame Interpolation},
  author = {Li, Zhen and Zhu, Zuo-Liang and Han, Ling-Hao and Hou, Qibin and Guo, Chun-Le and Cheng, Ming-Ming},
  booktitle = {Proceedings of the IEEE/CVF Conference on Computer Vision and Pattern Recognition},
  pages = {9801--9810},
  year = {2023}
}

@article{luo2026hifloat4,
  title = {{HiFloat4} Format for Language Model Inference},
  author = {Luo, Yuanyong and Huang, Jing and Cheng, Yu and Yu, Ziwei and Zhang, Kaihua and Hong, Kehong and Ma, Xinda and Wang, Xin and Tong, Anping and Hu, Guipeng and Xu, Yun and Taghian, Mehran and Wu, Peng and Li, Guanglin and Peng, Yunke and Hu, Tianchi and Chen, Minqi and Mi, Michael Bi and Liu, Hu and Zhou, Xiping and Wang, Junsong and Lin, Qiang and Liao, Heng},
  journal = {arXiv preprint arXiv:2602.11287},
  year = {2026},
  url = {https://arxiv.org/abs/2602.11287}
}

@article{yuan2025opens2v,
  title = {{OpenS2V-Nexus}: A Detailed Benchmark and Million-Scale Dataset for Subject-to-Video Generation},
  author = {Yuan, Shenghai and He, Xianyi and Deng, Yufan and Ye, Yang and Huang, Jinfa and Lin, Bin and Luo, Jiebo and Yuan, Li},
  journal = {arXiv preprint arXiv:2505.20292},
  year = {2025},
  url = {https://arxiv.org/abs/2505.20292}
}

\end{document}